\documentclass[3p,12pt]{elsarticle}
\usepackage{graphicx}
\usepackage{amssymb}
\usepackage{amsmath}
\usepackage{dsfont}
\usepackage[colorinlistoftodos]{todonotes}
\usepackage{threeparttable}
\usepackage{color,soul}
\usepackage{algorithm} 
\usepackage{algpseudocode}

\usepackage{amssymb}
\usepackage{makecell,multirow}
\usepackage[colorlinks=true]{hyperref}

\usepackage{geometry}
\usepackage{pdflscape}

\usepackage{soul}
\soulregister{\cite}7 
\soulregister{\citep}7 
\soulregister\citet7 
\soulregister\ref7 
\soulregister\pageref7 
\soulregister\title7


\newcommand{\review}[1]{{\color{black}{#1}}}

\begin{document}

\begin{frontmatter}
\title{Semantic-Fused Multi-Granularity Cross-City Traffic Prediction}

\author[1,2]{Kehua Chen}
\author[3]{Yuxuan Lian}
\author[1]{Jindong Han}
\author[5]{Siyuan Feng\corref{cor1}}
\author[3,4]{Meixin Zhu\corref{cor2}}
\author[2]{Hai Yang}

\address[1]{Division of Emerging Interdisciplinary Areas (EMIA), Academy of Interdisciplinary Studies, The Hong Kong University of Science and Technology, Hong Kong, China}
\address[2]{Department of Civil and Environmental Engineering, The Hong Kong University of Science and Technology, Hong Kong, China}
\address[3]{Intelligent Transportation Thrust, The Hong Kong University of Science and Technology (Guangzhou), Guangzhou, China}
\address[4]{Guangdong Provincial Key Lab of Integrated Communication, Sensing and Computation for Ubiquitous Internet of Things, Guangzhou, China}
\address[5]{Department of Logistics and Maritime Studies, The Hong Kong Polytechnic University, China}

\cortext[cor1]{sfengag@connect.ust.hk}
\cortext[cor2]{meixin@ust.hk}

\begin{abstract}
Accurate traffic prediction is essential for effective urban management and the improvement of transportation efficiency. Recently, data-driven traffic prediction methods have been widely adopted, with better performance than traditional approaches. However, they often require large amounts of data for effective training, which becomes challenging given the prevalence of data scarcity in regions with inadequate sensing infrastructures. To address this issue, we propose a Semantic-Fused Multi-Granularity Transfer Learning (SFMGTL) model to achieve knowledge transfer across cities with fused semantics at different granularities. In detail, we design a semantic fusion module to fuse various semantics while conserving static spatial dependencies via reconstruction losses. Then, a fused graph is constructed based on node features through graph structure learning. Afterwards, we implement hierarchical node clustering to generate graphs with different granularity. To extract feasible meta-knowledge, we further introduce common and private memories and obtain domain-invariant features via adversarial training. \review{It is worth noting that our work jointly addresses semantic fusion and multi-granularity issues in transfer learning. We conduct extensive experiments on six real-world datasets to verify the effectiveness of our SFMGTL model by comparing it with other state-of-the-art baselines. Afterwards, we also perform ablation and case studies, demonstrating that our model possesses substantially fewer parameters compared to baseline models. Moreover, we illustrate how knowledge transfer aids the model in accurately predicting demands, especially during peak hours. The codes can be found at \url{https://github.com/zeonchen/SFMGTL}.}
\end{abstract}

\begin{keyword}
Wastewater treatment \sep reinforcement learning \sep multi-objective optimization \sep sustainability


\end{keyword}
\end{frontmatter}


\section{Introduction}\label{Intro}
Traffic prediction is a critical concern in Intelligent Transportation Systems (ITS), it refers to the anticipation of future transportation conditions, encompassing factors such as traffic flow \citep{lv2014traffic,wang2022traffic}, traffic speed \citep{asif2013spatiotemporal,han2021dynamic}, origin-destination demand \citep{zhang2021dneat,shao2021deepflowgen,feng2021context}. The accurate prediction of these variables is instrumental in enhancing urban management and optimizing transportation efficiency, thereby alleviating traffic congestion, and reducing environmental impacts. 
This proliferation of data from diverse sources such as GPS devices, traffic cameras, and sensors integrated into vehicles and infrastructure, has changed the paradigm of traffic prediction.
Machine learning methods have emerged as powerful tools to harness massive data, these methods offer superior performance compared to traditional approaches \citep{tedjopurnomo2020survey}. 

However, the effective utilization of deep learning techniques often necessitates a substantial volume of data to attain desirable performance, and limited data size often leads to over-fitting problems. Given the data scarcity issue in cities with low developmental levels or new districts, it is non-trivial to train a powerful model. Actually, traffic patterns in different cities often exist common features. For instance, traffic speed in commercial areas decreases during peak time, and crowd flow from residential areas to commercial areas tends to surge in the morning. In light of these shared patterns, a method extracting traffic patterns from data-rich urban centers can be harnessed. Subsequently, we can adapt these acquired patterns to the context of data-scarce regions or cities. Such knowledge transfer among cities is called cross-city knowledge transfer and has attracted much attention these years \citep{chen2022cross}.

\begin{figure}[htbp]
    \centering    
    \includegraphics[width=15cm]{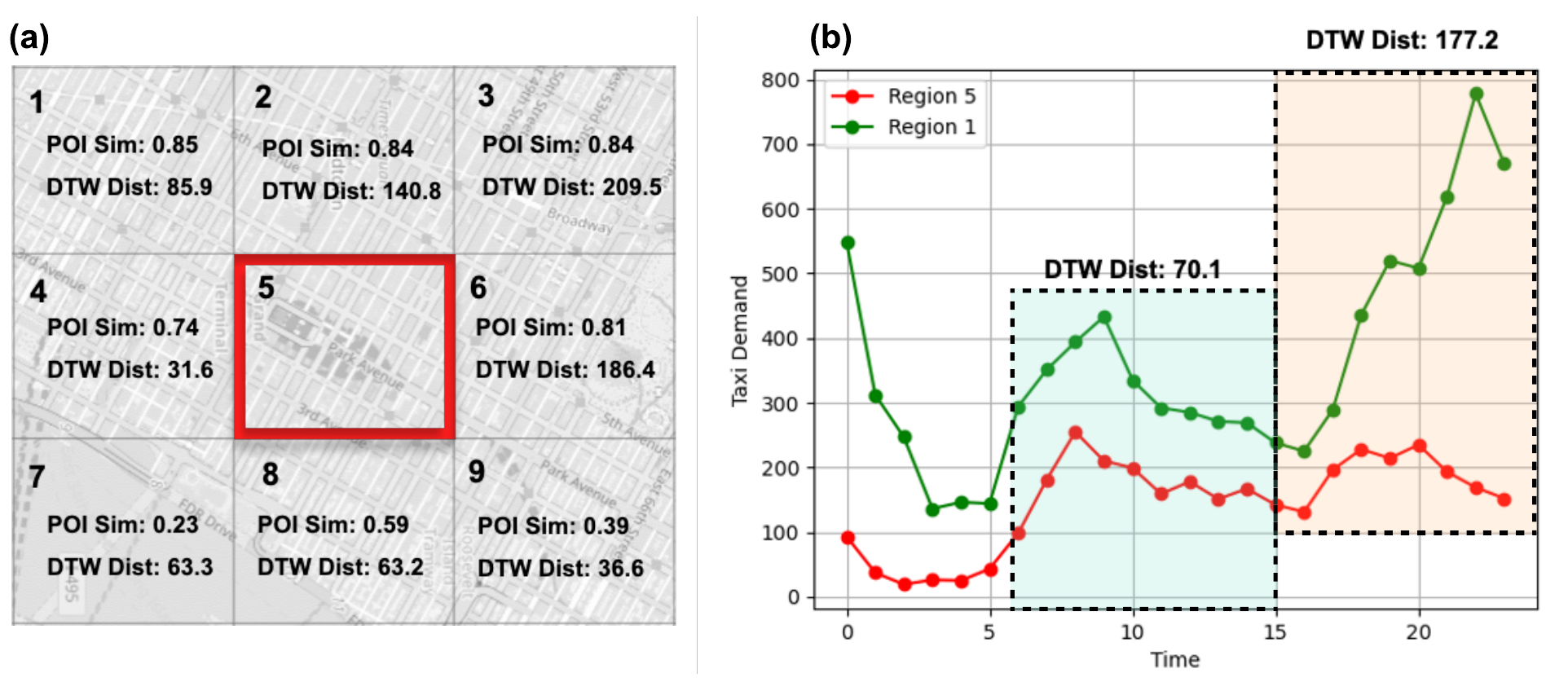}
    \caption{Current challenges in cross-city transfer learning. (a) presents the POI similarity and DTW distance among Region 5 and other regions; (b) demonstrates dynamic spatial dependencies.}
     \label{fig:intro}
\end{figure}

In terms of recent research, the constituents of cities such as road links, geographic regions, and subway stations are often conceptualized as nodes within a graph, and their distance or POI similarity can be regarded as edges. As such topological structures affect traffic patterns, there are still several challenges in graph-level knowledge transfer. \textbf{First}, the organization of urban data often involves the integration of diverse perspectives as multi-semantic adjacency matrices, discerned from long-term traffic datasets \citep{han2023kill,dai2021temporal,dai2023dynamic}. Previous studies either concatenated different semantics \citep{geng2019spatiotemporal} or applied weighted sum to fuse semantics \citep{chai2018bike}. 
However, it remains to be a non-trivial task to ascertain the advantageous impact of incorporating various perspectives on the prediction of traffic dynamics. For instance, Fig. \ref{fig:intro} (a) presents nine adjacent regions in New York City. The data showcases POI similarities and long-term Dynamic Time Warping (DTW) distances \citep{muller2007dynamic} (covering approximately one week) between Region 5 and its neighboring counterparts. Notably, Region 1 exhibits the highest POI similarity with Region 5, yet the temporal patterns of taxi demands in Region 5 exhibit greater similarity to those observed in Region 4 and Region 9 with lower DTW distances, despite their comparatively lower POI similarities. This incongruity demonstrates that the incorporation of multiple perspectives does not invariably confer favorable contributions to the predictive accuracy of the model.

\textbf{Second}, it is common to merely utilize short historical records to predict future traffic status under data-scarce scenarios. Nonetheless, it is imperative to acknowledge that the spatial interdependencies among regions tend to exhibit fluctuations over time. For example, Fig. \ref{fig:intro} (b) shows the DTW distances between Region 1 and 5 of short-term records within one day. Between 5 o'clock and 15 o'clock, the traffic fluctuations observed between these two regions align more closely with those exhibited between 15 o'clock and 24 o'clock. Such a dynamic pattern indicates that reliance on persistent historical dependencies might not necessarily yield performance gains for future predictions. The above two observations motivate us to introduce dynamic graph structure learning with fused semantics in the prediction model. 

\textbf{Third}, most previous studies merely achieved knowledge transfer at a solely local level without the consideration of coarse-scale knowledge. Recent studies such as \citet{mo2022cross} have proposed various methods to achieve multi-granularity transfer learning among cities. However, these studies either pre-established methods \citep{wang2022multivariate} for region clustering or solely focused on Euclidean relationships among regions.

To address the aforementioned challenges, we present an innovative framework aiming at tackling spatial-temporal transfer learning, denoted as Semantic-Fused Multi-Granularity Graph Transfer Learning (SFMGTL). In response to the first two challenges, we formulate a dedicated semantic fusion module, which acquires integrated node features while preserving the diversity of long-term semantics with a graph reconstruction process. To address the last challenge, we incorporate hierarchical node clustering, partitioning the graphs into distinct coarse zones for knowledge transfer. Additionally, we integrate the utilization of learnable meta-knowledge and implement adversarial training techniques to extract domain-invariant knowledge that further enriches the learning process. 

In summary, our contributions include the following aspects:
\begin{itemize}
    \item To the best of our knowledge, we present the first attempt to consider multi-semantic fusion in cross-city transfer learning. \review{The fused node features can adapt dynamic traffic status without compromising the preservation of static spatial relationships.}
    \item We employ a hierarchical graph clustering technique to establish coarse granularity exclusively through data-driven ways. Our model is trained using a strategy that involves simultaneous prediction of traffic status across varying scales. By adopting this approach, our model effectively incorporates and leverages multi-granular information, leading to a more comprehensive understanding of the underlying dynamics.
    \item \review{The extensive experiments demonstrate that our model outperforms state-of-the-art (SOTA) methods while owning significantly fewer model parameters. A further analysis of the quality of source datasets reveals that adding mild noise to source data degrades the average performance, but encourages the model to capture dynamic variations. Additionally, the case study illustrates that cross-city knowledge transfer primarily enhances predictions during peak hours.}
\end{itemize}

The remainder of the paper is organized as follows. We briefly introduce related works in Section 2, and propose definitions and problems in Section 3. Afterwards, the proposed model is revealed in Section 4. Section 5 thoroughly introduces experiments for evaluation, results and discussions. At last, we summarize the overall paper in Section 6.

\section{Related Work}
This paper is relevant to traffic prediction, transfer learning, and cross-city knowledge transfer. In this section, we briefly review related works for the above topics. 

\subsection{Traffic Prediction}
As a classical problem, researchers have focused on traffic prediction since several decades ago. Traditional statistical methods include Historical Average (HA) \citep{smith1997traffic}, Autoregressive Integrated Moving Average (ARIMA) \citep{van1996combining} and its variants. Nonetheless, these traditional methods merely capture linear relations and cannot perfectly depict data patterns. Hence, recent studies tend to use deep learning methods due to their powerful expressiveness. Researchers adopted various techniques to tackle traffic prediction problems, ranging from Recurrent Neural Network (RNN), Convolutional Neural Network (CNN) to Graph Neural Network (GNN). 

Long Short-Term Memory (LSTM) was applied in \citet{ma2015long} to predict traffic speed on a single road, and subsequent studies normally use RNN as a model component owing to its powerful capacity of capturing non-linear dynamics. 
Within a city, traffic status in one region is influenced by proximal regions. To model such relations, \citet{ma2017learning} formed traffic speed as various matrices, and then transformed matrices into image channels. Then CNN was applied to predict future traffic status. Yet, CNN can only capture Euclidean relations. As an emerging technique, GNN has been widely used in current models for traffic prediction. For example, \citet{li2017diffusion} modeled the traffic flow as a diffusion process on a directed graph and proposed DCRNN to forecast traffic speed. T-GCN \citep{zhao2019t} combined Graph Convolutional Network (GCN) and Gated Recurrent Unit (GRU) to predict traffic status. In addition, \citet{pan2019urban} came up with ST-MetaNet to capture traffic spatial and temporal correlations. The authors designed the model based on deep meta-learning and chose RNN and GNN as meta-learners.

Although deep learning methods achieve accurate traffic prediction, they typically require massive data. Hence, transfer learning provides an effective method to alleviate data scarcity.

\subsection{Transfer Learning}
Traditional machine learning approaches have achieved great success in various practical scenarios, such as object detection, and language processing. However, machine learning normally requires massive labeled data for training, and the collection of labeled data is time-consuming and expensive. 

Transfer learning achieves knowledge transfer from source domains to improve the learning performance in target domains. Hence, transfer learning is suitable to handle few-shot problems, i.e. rich data in source domains and scarce data in target domains.
In data-based interpretation \citep{zhuang2020comprehensive}, the main objective of transfer learning is to minimize the distribution difference between source and target domains, such as instance weighting strategy \citep{huang2006correcting} and feature transformation strategy \citep{pan2010domain}. In model-based interpretation \citep{zhuang2020comprehensive}, the aim is to accurately predict results on the target domain by utilizing source knowledge. As an emerging method, meta-learning \citep{vanschoren2018meta} has been used to achieve transfer learning as an effective method. In meta-learning, a number of tasks are drawn from task distributions, the model aims to learn a generalized meta-learner at the meta-training phase, and achieves fast adaption at the meta-testing phase when new tasks come. As one of the most famous methods, MAML \citep{finn2017model} treats the meta-learner as parameter initialization by bi-level optimization, we use MAML as the basic framework in this paper. Besides, \citet{achille2019task2vec} raised that the utilization of task discrepancy benefits the model performance, and various modulating methods have been applied in recent studies, such as automated relational meta-graph \citep{yao2020automated}, multimodal modulation network \citep{zhao2021multimodal}, and hierarchical prototype graph \citep{suo2020tadanet}.

While most previous transfer learning studies focused on image- and text-related tasks, spatial-temporal transfer learning is at the initial stage and still without much understanding. 

\subsection{Cross-City Knowledge Transfer}

To the best of our knowledge, \citet{wang2018cross} first proposed the Divide-Match-Transfer principle. The authors partitioned the cities into equal-size grid cells, then matched target regions with the most correlated source regions. At fine-tuning stage, the model tried to minimize the squared error between regional representations of target regions and matched source regions. 
MetaST \citep{yao2019learning} designed a memory module to extract long-term patterns from source cities. The regions in source cities were clustered into several categories as memory. Then, the attention mechanism was utilized to gather useful information from the memory during both the meta-training and meta-testing phases.
ST-DAAN \citep{wang2021spatio} first mapped source and target data to a common embedding space, and tried to minimize Maximum Mean Discrepancy (MMD) between two embeddings. To further capture spatial dependencies, the model introduced a global attention mechanism. The attention mechanism can also be deemed as a matching process. However, these studies utilized Euclidean relationships among regions and did not consider complex semantic information. 
\citet{lu2022spatio} designed the model in terms of graph data structure. The model extracted meta-knowledge through GRU and GAT. To express the structural information, graph construction loss was introduced. Last, the node-level meta-knowledge was fed into a parameter generation module to produce non-shared feature extractor parameters. \review{To dynamically adapt graph structures, TransGTR \citep{jin2023transferable} aimed to transfer graph structures among cities.}
To alleviate negative transfer, \citet{jin2022selective} proposed selective cross-city transfer learning to filter harmful source knowledge. The authors employed edge-level and node-level adaption to train the feature network, and designed a weighting network for loss calculation. \review{DastNet \citep{tang2022domain}, Ada-STGCN \citep{yao2023transfer} and CityTrans \citep{ouyang2023citytrans} employed adversarial training in the pre-training stage to extract common knowledge across cities. 
Although the above studies organized urban areas into graphs, cross-city knowledge transfer is conducted at the local level. }
For multi-granular transfer learning, MGAT \citep{mo2022cross} utilized multiple convolutional kernels to extract multi-granular information, then applied an attention mechanism to share knowledge across cities. However, they solely focused on Euclidean relationships among regions.

\section{Problem Formulation}
In this section, we introduce the notations and formally define the cross-city knowledge transfer problem.

\noindent\textit{Definition 1 (Graph):} Given a graph with $C$ different semantics (or views) as $\mathcal{G}=(\mathcal{V}, \{\mathcal{E}_c\}_{c=1}^C, \textbf{Z}, \{\textbf{A}_c\}_{c=1}^C)$. $\mathcal{V}=\{v_1, v_2, ...., v_N\}$ is the node set, and $N=|\mathcal{V}|$; $\mathcal{E}_c=\{e_{ij}^c=(v_i, v_j)\}$ is the edge set of semantic $c$; $\textbf{Z}\in \mathds{R}^{N\times D}$ is the node feature matrix, and the feature dimension of each node is $D$; $\textbf{A}_c$ is the adjacency matrix under view $c$, each entry $a_{ij}^c$ indicates the weight of edge $e_{ij}^c$. 



\noindent\textit{Problem Statement 1 (Traffic Prediction):} Consider there are $N$ sensors/recorders in a given city area, sensor/recorder $n$ measures $F$ traffic status on a specific node (e.g. road link, or region) at time $t$, denoted as $\bold X_t\in \mathds{R}^{N\times T\times F}$. Given past $T$ records and corresponding graph structures, the sequential traffic status can be naturally deemed as node features, then traffic prediction aims to find a function $g(\cdot)$ that predicts the next $M$ traffic status:
\begin{align}
\nonumber
    \{\bold X_{t-T+1},...,\bold X_t, \bold A_1,...,\bold A_C\}\stackrel{g(\cdot)}{\longrightarrow}
    \{\bold X_{t+1}, ..., \bold X_{t+M}\}
\end{align}
For convenience, we denote the input and the corresponding label pair as $(\mathcal{X}, \mathcal{Y})$.

\begin{figure}[t]
    \centering    
    \includegraphics[width=17cm]{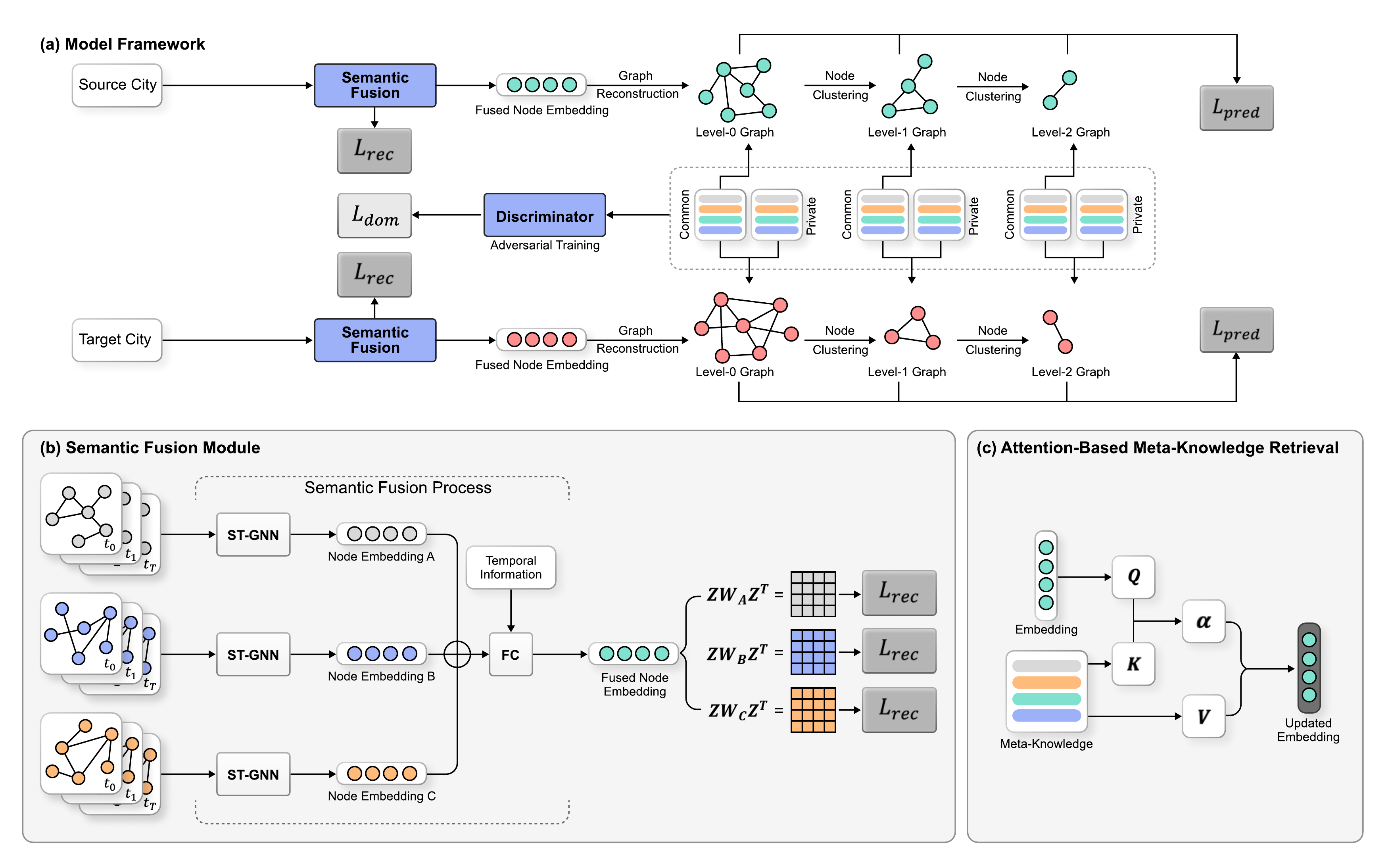}
    \caption{\review{The proposed model structure.}}
     \label{fig:model}
\end{figure}

\noindent\textit{Problem Statement 2 (Cross-City Traffic Prediction):} Here we assume only one source and one target city. Given a source city denoted as $\mathcal{C}_\mathcal{S}$ and a target city denoted as $\mathcal{C}_\mathcal{T}$. The source city is data-rich while the target city is data-scarce. \review{We aim to find a function $f(\cdot)$ parameterized by $\theta$ that leverages knowledge of the source city to predict traffic status in the target city as accurately as possible:}

\review{
\begin{align}
&\theta_\mathcal{S} = \text{arg min}_\theta \mathcal{L}_\mathcal{S}(\mathcal{C}_\mathcal{S},\theta)  \\
&\theta_\mathcal{T} = \text{arg min}_\theta \mathcal{L}_\mathcal{T}(\mathcal{C}_\mathcal{T},\theta;\theta_{\mathcal{S}})
\end{align}
where Eq. (1) is the source training, and Eq. (2) is the fine-tuning stage based on target datasets; $\mathcal{L}_{\mathcal{S}}(\cdot)$ and $\mathcal{L}_{\mathcal{T}}(\cdot)$ are loss functions of two stages.}

\section{Methodology}
Fig. \ref{fig:model} (a) illustrates the proposed model structure. The training of our model includes two stages. In the pre-training stage, the historical traffic data and the corresponding adjacency matrices pertaining to both the source and target urban areas are utilized as inputs for the semantic fusion modules. This process leads to the generation of fused node embeddings. Subsequently, graph reconstruction is executed, followed by the implementation of hierarchical node clustering. In order to distill meta-knowledge of dynamic traffic patterns, the model incorporates learnable common memories and exclusive private memories that are tailored to the target city, each functioning at a distinct level of granularity. 
The node embeddings are deemed as queries to retrieve meta-knowledge via the attention mechanism. Additionally, we employ adversarial training to ensure the learned meta-knowledge is domain-invariant. Last, the model is trained to predict future traffic states across varying levels of granularity in a concurrent manner. At the fine-tuning stage, we freeze the learned common knowledge, while the training exclusively focuses on data originating from the target city.

\subsection{Semantic Fusion}
Fig. \ref{fig:model} (b) presents the semantic fusion module. As aforementioned, we aim to learn dynamic node embeddings by leveraging historical short-term traffic conditions, while simultaneously upholding long-term spatial interdependencies. Given the presence of $C$ distinct semantics, the initial step involves the feeding of sequential graphs specific to each semantic into Spatial-Temporal Graph Neural Networks (ST-GNN). This procedure leads to the extraction of semantic-oriented node embeddings denoted as $\bold{Z}_c$. Please note that any off-the-shelf ST-GNN such as T-GCN \citep{zhao2019t} and DCRNN \citep{li2017diffusion} can be employed for this purpose.

Subsequently, the extracted node embeddings are concatenated, and a fused node embedding is generated through the utilization of a fully connected neural network (FC). To further consider the temporal difference, we also input temporal information $\bold T \in\mathds{R}^{7+24}$ including day of the week and hour of the day together into the FC:
\begin{align}
    \bold Z_f = \text{FC}(\bold Z_1\oplus...\oplus\bold Z_C\oplus \bold T)
\end{align}

Based on the fused node embedding, we can generate a fused adjacency matrix $\bold A_f$ as a momentary graph similar to \citet{zhang2020spatio}:
\begin{align}
    \bold A_f = \text{Softmax}(\text{ReLU}((\bold Z_f\bold W_f)(\bold Z_f\bold W_f)^\top))
\end{align}
where $\bold W_f$ is the learnable parameters.

In addition, we present a reconstruction process aimed at preserving diverse static urban structures. Specifically, our approach involves reconstructing the adjacency matrix $\bold A_c$ for each semantic via the fused node embeddings:
\begin{align}
    \hat{\bold A}_c = \bold Z_f\bold W_c\bold Z_f^\top
\end{align}
where $\bold W_c$ is the learnable parameters for semantic $c$.

Then we acquire the reconstruction loss as:
\begin{align}
    \mathcal{L}_{rec} = \frac{1}{C}\sum_{c=1}^C||\bold A_c - \hat{\bold A}_c||_2^2
\end{align}

\review{To the best of our knowledge, our SFMGTL first dynamically fuses multiple semantics in cross-city transfer learning. Compared with previous studies, our semantic fusion procedure demonstrates a notable capability in capturing the dynamic characteristics inherent in node embeddings rather than simply combining existing adjacency matrices.}

\subsection{Hierarchical Node Clustering}
\begin{figure}[h]
    \centering    
    \includegraphics[width=8cm]{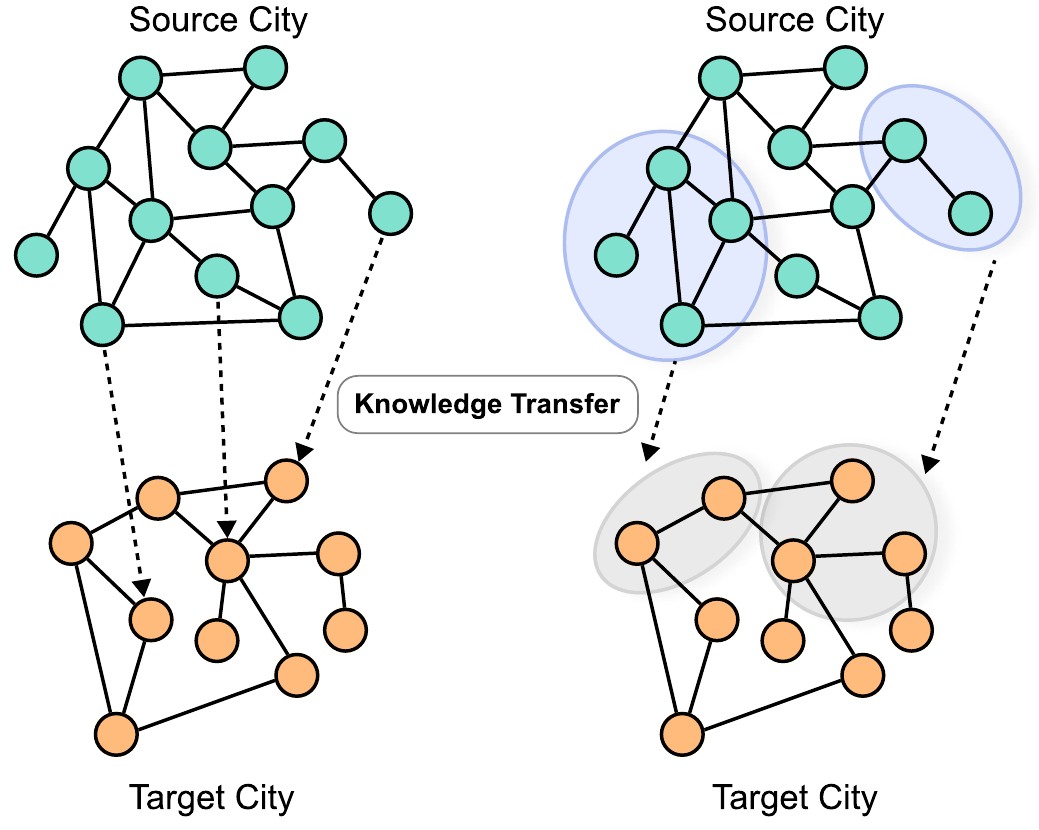}
    \caption{Multi-granularity knowledge transfer.}
     \label{fig:graular}
\end{figure}

Urban areas with multi-granularity can provide various information \citep{qiao2020tree,han2022continuous}. Inspired by \citet{ying2018hierarchical}, our approach acquires soft assignment matrices from the data, enabling the aggregation of nodes as illustrated in Fig. \ref{fig:graular}. Denote the input node embeddings at layer $l$ as $\bold Z^{(l)}_f\in \mathds{R}^{N_l\times D}$ with $N_l$ nodes, the learned soft assignment matrix at layer $l$ as $\bold S^{(l)}\in \mathds{R}^{N_l\times N_{l+1}}$, the fused adjacency matrix as $\bold A^{(l)}_f\in \mathds{R}^{N_{l}\times N_l}$. Consequently, our framework facilitates the aggregation of information from finer to coarser granularities:
\begin{align}
&\bold S^{(l)} = \text{Softmax}(\text{GNN}^{(l)}(\bold A^{(l)}_f, \bold{Z}^{(l)}_f))\\
&\bold{Z}^{(l+1)}_f = \bold S^{{(l)}^\top}\bold Z^{(l)}_f\\
&\bold A^{(l+1)}_f = \bold S^{(l)^\top}\bold A^{(l)}_f\bold S^{(l)}
\end{align}
where $\text{GNN}^{(l)}$ is a Graph Neural Network of level $l$.

After acquiring soft assignment matrices at each aggregation level, we further generate labels for each granularity:

\begin{align}
    \mathcal{Y}^{(l+1)} = \bold S^{{(l)}^\top}\mathcal{Y}^{(l)}
\end{align}

Then we can deduce the prediction $\hat{\mathcal{Y}}^{(l)}$ based on the attention mechanism and meta-knowledge as introduced in the next subsection. Our model simultaneously predicts traffic status across various levels of granularity in a multi-task fashion:

\review{
\begin{align}
    \mathcal{L}_{pred}^{(l)} = ||\mathcal{Y}^{(l)}-\hat{\mathcal{Y}}^{(l)}||_2^2
\end{align}}

Besides, we incorporate a hierarchical link prediction loss aimed at capturing the interdependence of neighboring nodes, and hierarchical entropy loss to force exclusive assignment in line with the methodology established in \citet{ying2018hierarchical}:

\begin{align}
    &\mathcal{L}_{aux} = ||\bold A^{(l)}_f, \bold S^{(l)} \bold S^{(l)^\top}||_F + \frac{1}{N_l}\sum_{i=1}^{N_l}H(\bold S^{(l)}(i))
\end{align}
\review{where $||\cdot||_F$ is the Frobenius norm; $H(\cdot)$ means the entropy function; $\bold A_f^{(l)}$ indicates the fused adjacency matrix at layer $l$; and $\bold S^{(l)}(i)$ is the $i$-the row of $\bold S^{(l)}$. The first term is also known as the auxiliary link prediction term, ensuring soft-assignment matrices encode nearby nodes closely. It should be noted that the original node clustering model in \citet{ying2018hierarchical} aims to generate graph embeddings, whereas our model utilizes it to extract multi-granular regional information.}

\subsection{Domain-Invariant Meta-Knowledge Memory}
Domain shift is ubiquitous in transfer learning, and the inappropriate transference of knowledge across domains can result in detrimental negative transfer effects. Hence, it is crucial to extract domain-invariant common knowledge during learning. In this study, we adopt memory modules akin to those presented in \citet{yao2019learning} and integrate an adversarial training process to learn domain-invariant common knowledge as Fig. \ref{fig:model} (a) and (c) present. 

Specifically, we establish learnable domain-invariant knowledge at each level as $\bold I^{(l)} \in \mathds{R}^{M\times D}$, with $M$ being a hyperparameter determining the number of learnable features. Afterwards, we treat the node embeddings $\bold Z_f^{(l)}$ as queries $\bold Q^{(l)}$ to retrieve meta-knowledge via attention mechanism:
\begin{align}
    &\bold Q^{(l)} = \bold Z_f^{(l)}\bold W_q, \bold K^{(l)} = \bold I^{(l)}\bold W_k, \bold V^{(l)} = \bold I^{(l)}\bold W_v\\
    &\alpha(\bold Q^{(l)}(i), \bold K^{(l)}(j)) = \frac{\text{exp}(\bold Q^{(l)^\top}(i)\bold K^{(l)}(j))}{\sum_{i'=1}^{M}\text{exp}(\bold Q^{(l)}(i')^\top\bold K^{(l)}(j))}\\
    &\bold O^{(l)}(i) = \sum_{j=1}^M\alpha(\bold Q^{(l)}(i), \bold K^{(l)}(j))\bold V^{(l)}(j)
\end{align}
where $\bold W_q, \bold W_k, \bold W_v\in \mathds{R}^{D\times D}$ are learnable parameters; $\bold O^{(l)}$ is the updated node embedding at layer $l$.

Since our target is to learn an adaptive model in the target city, we extend our approach by incorporating distinct private memories to house domain-specific information unique to the target city. Hence, compared to the memories of source cities, the number of learnable features of the target city exceeds $M$. And we denote the updated node embeddings as $\bold O_S^{(l)}$ and $\bold O_T^{(l)}$ for source and target cities respectively. For single-step prediction, we employ simple FC layers as the prediction head to generate predictions for each node:
\begin{align}
    \mathcal{Y}^{(l)} = \text{FC}(\bold O^{(l)}_{(\cdot)})
\end{align}
Certainly, other sequential decoders \citep{zheng2020gman,du2019lstm} can be used for multi-step prediction.

Afterwards, to ensure the learned common knowledge is domain-invariant, we employ adversarial training between source and target cities similar to \citet{tang2022domain} and \citet{tuli2022tranad}. In detail, we use a discriminator $\mathcal{D}$ consisting of multiple fully connected layers to distinguish queries $\bold Q^{(l)}$ from source and target cities, while the hierarchical node clustering encoders try to fool the discriminator, thereby establishing a competitive dynamic:
\begin{align}
    \mathcal{L}_{dom}^{(l)} = - \frac{1}{N_l}\sum_{i=1}^{N_l}\text{log}\mathcal{D}(\bold Q^{(l)}_S) - \frac{1}{N_l}\sum_{i=1}^{N_l}\text{log}[1 - \mathcal{D}(\bold Q^{(l)}_T)]
\end{align}

\subsection{Training Process}
For either source or target cities, we have the following overall loss function:
\begin{align}
    \mathcal{L}_{(\cdot)} = \sum_{l=1}^L\mathcal{L}_{pred}^{(l)} + \lambda_1\cdot \mathcal{L}_{rec} + \lambda_2\cdot \mathcal{L}_{aux}
\end{align}
where $\lambda_1, \lambda_2$ are hyperparameters to balance the overall loss function; and $\mathcal{L}_{(\cdot)}$ can be the overall loss function of source or target cities. 

During the pre-training process as shown in Algorithm \ref{algo1}, we first train the source and target city together but emphasize the performance of the source city:
\begin{align}
    \mathcal{L} = \mathcal{L}_S + \beta_1\cdot \mathcal{L}_T + \beta_2\cdot \sum_{l=1}^L\mathcal{L}_{dom}^{(l)}
\end{align}
where $\beta_1, \beta_2$ are two hyperparameters.

In the fine-tuning stage as shown in Algorithm \ref{algo2}, we freeze the parameters of common memories, and then solely train on datasets from the target city based on $\mathcal{L}_T$.

\begin{algorithm}
    \small
    \caption{SFMGTL pre-training algorithm}
    \label{algo1}   
    \begin{algorithmic}[1]
    \Require Source and target city datasets $\mathcal{C}_s$ and $\mathcal{C}_T$.
	\Ensure Pre-trained model for target city.
	\State Randomly initialize model parameters.
	\While{\textit{not done}}
        \State Sample a batch of data from $\mathcal{C}_S$ and $\mathcal{C}_T$.
        \State For either $\mathcal{C}_S$ and $\mathcal{C}_T$, generate $\{\bold Z_c\}_{c=1}^C$ for all semantics via specific ST-GNN.
        \State Generate node embeddings $\bold Z_f$ for cities via Eq. (1).
        \State Generate adjacency matrices $\bold A_f$ for cities via Eq. (2).
        \State Calculate the reconstruction loss for cities via Eq. (4).
        \State Implement hierarchical node clustering for both cities via Eq. (5)-(7).
        \State Obtain labels at each granularity via Eq. (8).
        \State Calculate auxiliary losses via Eq. (10).
        \State Predict traffic status at each granularity by attention mechanism via Eq. (11)-(14).
        \State Calculate prediction loss and adversarial loss via Eq. (9) and Eq. (15).
        \State Update the model parameters via Eq. (16)-(17).
	\EndWhile
    \end{algorithmic}
\end{algorithm}

\begin{algorithm}
    \small
    \caption{SFMGTL fine-tuning algorithm}
    \label{algo2}   
    \begin{algorithmic}[1]
    \Require Pre-trained model and target city dataset $\mathcal{C}_T$.
	\While{\textit{not done}}
        \State Sample a batch of data from $\mathcal{C}_T$.
        \State Generate $\{\bold Z_c\}_{c=1}^C$ for all semantics via specific ST-GNN.
        \State Generate node embeddings $\bold Z_f$ via Eq. (1).
        \State Generate adjacency matrices $\bold A_f$ via Eq. (2).
        \State Calculate the reconstruction loss via Eq. (4).
        \State Implement hierarchical node clustering via Eq. (5)-(7).
        \State Obtain labels at each granularity via Eq. (8).
        \State Calculate auxiliary losses via Eq. (10).
        \State Predict traffic status at each granularity by attention mechanism via Eq. (11)-(14).
        \State Calculate prediction loss and via Eq. (9).
        \State Update the model parameters via Eq. (16).
	\EndWhile
	
    \end{algorithmic}
\end{algorithm}

\section{Experiment}
To evaluate the effectiveness of our framework, we implement several extensive experiments with six real-world datasets and compare the model performance with multiple baseline methods. The experiments are related to taxi and bike demand prediction in various cities. Importantly, our framework's adaptability extends to other traffic prediction tasks, including but not limited to traffic speed and flow prediction. Furthermore, we undertake ablation and sensitivity analysis to delineate the significance of each constituent within our model and the sensitivity of various hyperparameters. 
\review{Additionally, we conduct a comparative analysis based on varying qualities of source datasets and present a case study to further illustrate our findings.} Such in-depth analyses not only serve to highlight the robustness of our model but also offer insights into the optimal configurations.

\subsection{Dataset}
We verify our model by predicting taxi and bike pickup/dropoff demands in different regions. The transfer learning involves six datasets following \citet{jin2022selective} and \citet{mo2022cross}: NY-Taxi/Bike, CHI-Taxi/Bike, and DC-Taxi/Bike. We divide the study area into various 1 km$\times$1 km grid cells and time interval of 1 hour. In this study, we introduce three long-term urban graphs from different views: proximity graph, road connection graph and POI graph. These graphs provide diverse and complementary insights of the urban regions, enabling our model to capture the intricate spatial-temporal relationships of transportation demand. Note that our method can be easily extended to other semantic graphs:
\begin{itemize}
    \item \textbf{Proximity graph}: we link each grid cell in the city with 8 neighboring grid cells to indicate the adjacent relations.
    \item \textbf{Road connection graph}: the weights are determined by the number of connected highways.
    \item \textbf{POI graph}: we employ cosine similarity to calculate the POI similarity between two regions. The POI categories are the same as introduced in \citet{jin2022selective}.
\end{itemize}

During the prediction phase, we utilize the past 6 time steps to forecast the next time step following previous studies \citep{jin2022selective}. The preprocessing stage involves the application of Min-Max normalization, ensuring that the model operates effectively and consistently across different datasets and variables. \review{The datasets from the source cities cover a 12-month period, which is fully utilized for training purposes. However, the training sets from the target city are constrained, specifically limited to 1-day, 3-day, and 7-day intervals in this study. For validation and testing, the last 4 months of data are employed, with two months allocated to each phase. } Table \ref{Tab:description} presents the statistics of the datasets.

\begin{table}[h]
\caption{The description and statistics of datasets}
\centering
\begin{tabular}{cccc}
\hline
\textbf{City} & \textbf{New York}    & \textbf{Chicago}     & \textbf{Washington}  \\
\hline
Latitude  & 40.65-40.85 & 41.77-42.01 & 38.80-38.97 \\
Longitude & 74.06-73.86 & 87.74-87.58 & 77.13-76.93 \\
Time      & \multicolumn{3}{c}{1/1-12/31, 2016}     \\
\# Taxi   & 134M        & 25M         & 10M         \\
\# Bike   & 14M         & 4M          & 3M      \\   
\hline
\end{tabular}
\label{Tab:description}
\end{table}

\subsection{Experiment Setup}

Since the prediction problem is a regression problem, in this study, \review{we utilize Mean Average Error (MAE), Root Mean Square Error (RMSE) and R$^2$ score as metrics.} Since there exists plenty of zero values in the study areas, we do not employ Mean Absolute Percentage Error (MAPE) in this study.
\begin{align}
    &\text{RMSE} = \sqrt{\frac{1}{N}\sum_{i=1}^N(y_i-\hat{y}_i)^2}\\
    &\text{MAE} = \frac{1}{N}\sum_{i=1}^N|y_i-\hat{y}_i|\\
    &\review{\text{R}^2=1-\frac{\sum_{i=1}^N(y_i-\hat{y}_i)^2}{\sum_{i=1}^N(y_i-\bar{y})^2}}
\end{align}
where $\hat{y}_i$ and $y_i$ are predicted and true values of instance $i$ respectively; \review{$\bar{y}$ is the mean value of true values}.

The experiments are conducted on a Linux server with one NVIDIA RTX 3090 GPU. We adopt T-GCN \citep{zhao2019t} as the ST-GNN model to capture both temporal and spatial relations. In addition, we use Graph Attention Networks (GAT) \citep{velickovic2017graph} with 2 heads to process graph data structure in a hierarchical node clustering process. The node embeddings are derived via mean pooling. The hidden size of SFMGTL is 32, and the number of learnable features for both common and private memories is 3. We implement hierarchical node clustering twice with the numbers of nodes 100 and 10 respectively. For FC layers in the semantic fusion module and prediction heads, they consist of two linear layers with LeakyReLU as the activation function. As for adversarial training, the discriminator also includes two linear layers connecting with LeakyReLU function. We employ Gradient Reversal Layer (GRL) \citep{ganin2015unsupervised} to reverse the gradient of the generator. And we let $\lambda_1=\lambda_2=1.0$, $\beta_1=0.5$ and $\beta_2=1.0$ respectively. The learning rate is 1e-3 and Adam \citep{kingma2014adam} is employed as the optimizer. Additionally, to simulate scenarios with limited data availability, we utilize 1-day, 3-day, and 7-day datasets within the target cities for training purposes. We conducted three parallel experiments for comparison.
\vspace{0.3em}

\subsection{Baselines}
To verify the model effectiveness, we utilize 14 baselines for comparison, including traditional and state-of-the-art cross-city transfer learning methods.

\begin{itemize}
    \item ARIMA \citep{hyndman2018forecasting}: Autoregressive Integrated Moving Average is a classic method for time-series prediction and merely considers linear relations among data.
    \item GRU \citep{cho2014properties}: Gated Recurrent Unit is a popular deep learning method to capture sequential dependency of data.
    \item ST-Net \citep{yao2019learning}: ST-Net combines CNN and RNN together to handle spatial-temporal variations.
\end{itemize}

The above methods merely utilize the data of target cities for model learning and do not involve the knowledge transfer process. Next, we pick several state-of-the-art transfer learning models for comparison:
\begin{itemize}
    \item Fine-tuned Model: As a natural idea, we first train the base learner through source cities, and then fine-tune the model based on target cities.
    \item MAML \citep{finn2017model}: As mentioned before, MAML uses bi-level optimization for parameter initialization, and the base learner is TGCN.
    \item ST-DAAN \citep{wang2021spatio}: ST-DAAN employs ConvLSTM to extract spatial-temporal features from the data. Then the knowledge is transferred by minimizing Maximum Mean Discrepancy. 
    \item DastNet \citep{tang2022domain}: DastNet utilizes adversarial training during the pre-training stage to learn domain-invariant representations from limited data. 
    \item MGAT \citep{mo2022cross}: MGAT extracts multi-granular information based on various convolutional kernels, and generates multi-granular meta-knowledge for fast adaption.    
    \item MetaST \citep{yao2019learning}: MetaST is based on MAML framework, and it introduces a memory module to store common functional information.
    \item ST-GFSL \citep{lu2022spatio}: ST-GFSL generates non-shared parameters based on node-level meta-knowledge. Parameter matching is implemented to retrieve similar features from source cities.
    \item CrossTReS \citep{jin2022selective}: CrossTReS proposed a selective transfer learning framework to pick beneficial knowledge from the source domain by introducing a weighting network. Besides, it is suitable to handle multi-graph spatial-temporal prediction problems.
    \item \review{TransGTR \citep{jin2023transferable}: TransGTR is a transferable structure learning framework for traffic forecasting. It employs a graph structure generator to dynamically update adjacency matrices.}
    \item \review{Ada-STGCN \citep{yao2023transfer}: Ada-STGCN employs a Spatial-Temporal Graph Convolutional Network to extract features from both source and target domains. The authors also applied the adversarial domain adaptation to learn domain-invariant knowledge.}
    \item \review{CityTrans \citep{ouyang2023citytrans}: CityTrans consists of a spatial-temporal network, the domain-adversarial training strategy, and a knowledge attention mechanism without two-stage training.}
\end{itemize}

\subsection{Overall Results}
\begin{figure}[htbp]
    \centering    
    \includegraphics[width=10cm]{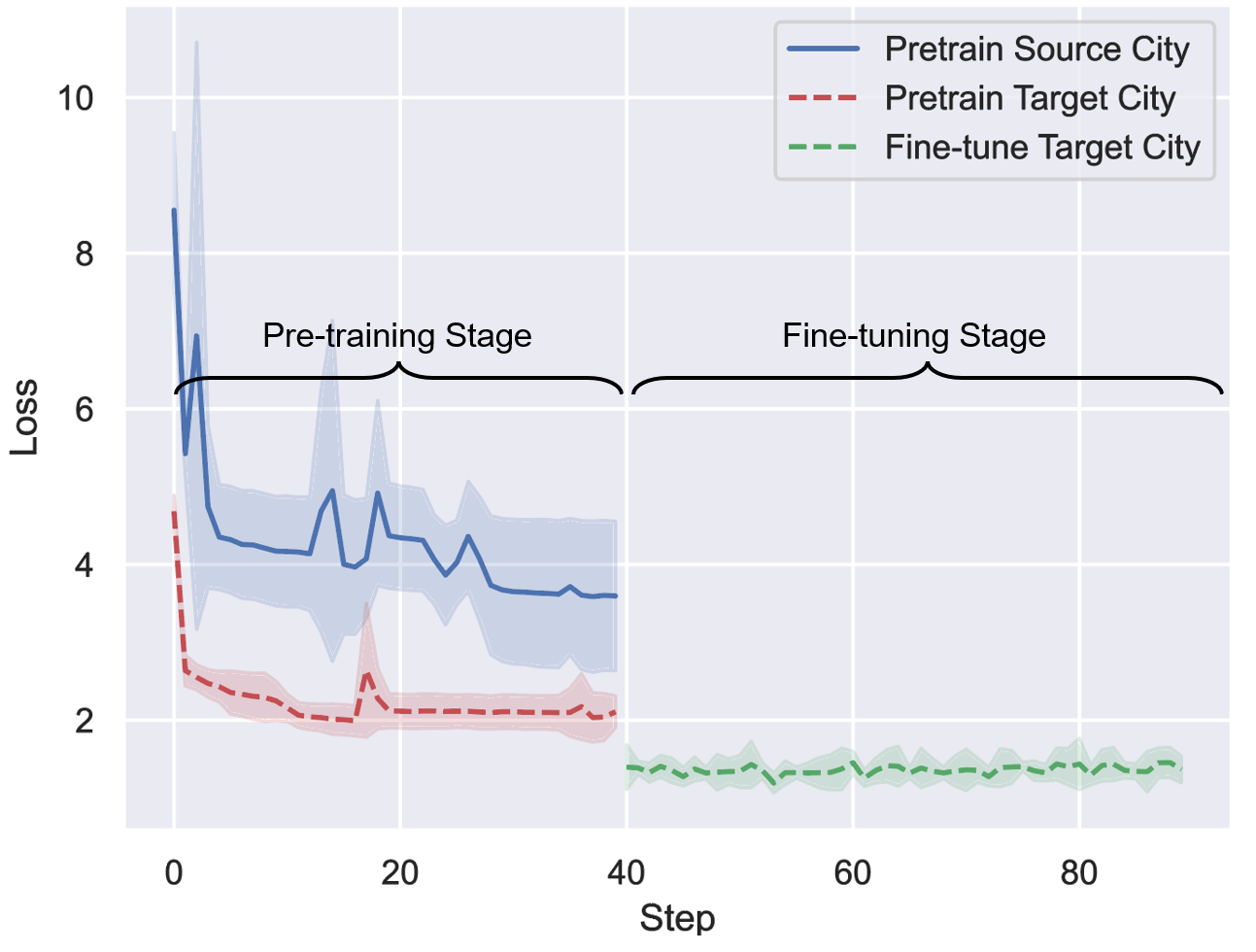}
    \caption{Training loss during pre-training and fine-tuning stages.}
     \label{fig:train}
\end{figure}

Fig. \ref{fig:train} presents the variation of losses during pre-training and fine-tuning. The training losses of both source and target cities decrease gradually in the pre-training stage and the loss of the target city further reduces in the fine-tuning stage. Table \ref{Tab:res1} and \ref{Tab:res2} shows the experimental results of taxi/bike demand prediction. The best result is bold, and the second-best result is underlined. 
Notably, deep learning methods exhibit a notable superiority over conventional techniques since they can capture complicated non-linear spatial-temporal relationships. Benefiting from knowledge sharing, this advantage is further enhanced when employing transfer-based techniques, resulting in an augmentation of model performance. Based on the results, it becomes evident that both fine-tuning methods and models based on the MAML framework yield commendable performance. 

\review{Table \ref{Tab:compare} provides a comparison of the number of parameters along with the average improvements of SFMGTL relative to each model. 
The performance superiority of SFMGTL over SOTA models by 2-4\% can be attributed to its integration of semantic fusion and multi-granularity knowledge transfer mechanisms, combined with the utilization of adversarial training to mitigate negative transfer effects.
Furthermore, CrossTReS achieves the second-best performance in most cases since only CrossTReS and our SFMGRL consider employing multiple semantics for prediction. In contrast, recent models such as TransGTR, Ada-STGCN, and CityTrans still rely on predictions based on a single adjacency matrix.
Furthermore, SFMGTL achieves these performance enhancements while maintaining a significantly lower number of parameters (i.e. 114.8 K) compared to CrossTReS with 482.7 K. It should be noted that the rise of model parameters not only elongates inference time and enlarges model sizes but also increases training complexity by potentially leading to overfitting issues.
}
Additionally, the performance of ST-DAAN and MetaST seems to lag behind that of more recent approaches. The main reason is that ST-DAAN and Meta-ST neither consider underlying graph structures and multi-granularity, nor integrate specialized modules to counteract the adverse effects of negative transfer in intercity scenarios. 

\review{We also provide the training results of SFMGTL using the full-scale dataset of the target cities (i.e., 8 months). 
Given the sufficient data size in this setting, no transfer learning is employed.
Compared to the full-scale dataset (denoted as SFMGTL-Full), limited data leads to a decrease in prediction accuracy by around 10-30\% in terms of MAE for SFMGTL. 
As SFMGTL achieves the best performance in the overall results, this observation further underscores that data scarcity significantly harms traffic prediction accuracy, and transfer learning can mitigate this issue.}

\begin{table}[htbp]
\scriptsize
\renewcommand\arraystretch{1}
\caption{The experimental results for taxi demand prediction}
\centering
\review{\begin{tabular}{c|c|c|c|c|c|c|c|c|c}
\hline
\multirow{3}*{\textbf{Model}}&\multicolumn{9}{c}{\textbf{NY-Taxi}$\xrightarrow{}$ \textbf{DC-Taxi}}\\
\cline{2-10}
&\multicolumn{3}{c}\textbf{RMSE\quad\quad\quad\quad\quad}&\multicolumn{3}{|c}\textbf{MAE\quad\quad\quad\quad\quad}&\multicolumn{3}{|c}\textbf{$\text{R}^2$\quad\quad\quad\quad\quad}\\
\cline{2-10}
&\textbf{1 Day}&\textbf{3 Day}&\textbf{7 Day}&\textbf{1 Day}&\textbf{3 Day}&\textbf{7 Day}&\textbf{1 Day}&\textbf{3 Day}&\textbf{7 Day}\\
\hline
\(\textbf{ARIMA}\) &6.842&6.547&6.464 &2.235&2.014 &1.968&0.839 &0.852 &0.856 \\
\(\textbf{GRU}\) & 5.522&4.714&4.588&2.282&1.737&1.632&0.895 &0.923 &0.927 \\
\(\textbf{ST-Net}\) &6.251 &5.045 &4.934 &1.985 &1.721 &1.703&0.865 &0.912 &0.916 \\
\(\textbf{Fine-tuning}\) &5.140 &4.780&4.367&1.719&1.520&1.517&0.909 &0.921 &0.934 \\
\(\textbf{MAML}\) &5.909&4.644&4.506&1.849&1.561&1.464&0.880 &0.926 &0.930 \\
\(\textbf{DastNet}\)&5.185&4.375&4.119&1.651&1.529&1.439&0.907 &0.934 &0.942 \\
\(\textbf{MGAT}\)&5.134&\underline{4.372}&4.093&1.646&1.520&1.439& 0.909 &\underline{0.934} &0.942 \\
\(\textbf{ST-DAAN}\) &5.240&4.419&4.147&1.678&1.522&1.435&0.905 &0.933 &0.941 \\
\(\textbf{MetaST}\) &5.221&4.413&4.145&1.669&1.516&1.442&0.906 &0.933 &0.941 \\
\(\textbf{ST-GFSL}\) & 5.106&{4.376}&\underline{4.058}&1.684&\underline{1.506}&1.431&0.910 &0.934 &\underline{0.943} \\
\(\textbf{CrossTReS}\) &\underline{5.039}&4.381&{4.089}&\underline{1.674}&1.584& \underline{1.429}&\underline{0.912} &0.934 &0.942 \\
\(\textbf{TransGTR}\)& 5.108&4.491&4.143&1.700&1.522&1.467&0.910 &0.930 &0.941 \\
\(\textbf{Ada-STGCN}\)&5.140&4.370&4.094&1.676&1.527&1.433&0.909 &0.934 &0.942 \\
\(\textbf{CityTrans}\)&5.165&4.366&4.105&1.687&1.535&1.463&0.908 &0.934 &0.942 \\
\hline
\(\textbf{SFMGTL}\)&\textbf{5.003}&\textbf{4.331} &\textbf{4.025} &\textbf{1.617} &\textbf{1.503} &\textbf{1.425}&\textbf{0.914} &\textbf{0.935} &\textbf{0.944} \\
\hline
\multirow{3}*{\textbf{Model}}&\multicolumn{9}{c}{\textbf{CHI-Taxi}$\xrightarrow{}$\textbf{DC-Taxi}}\\
\cline{2-10}
&\multicolumn{3}{c}\textbf{RMSE\quad\quad\quad\quad\quad}&\multicolumn{3}{|c}\textbf{MAE\quad\quad\quad\quad\quad}&\multicolumn{3}{|c}\textbf{$\text{R}^2$\quad\quad\quad\quad\quad}\\
\cline{2-10}
&\textbf{1 Day}&\textbf{3 Day}&\textbf{7 Day}&\textbf{1 Day}&\textbf{3 Day}&\textbf{7 Day}&\textbf{1 Day}&\textbf{3 Day}&\textbf{7 Day}\\
\hline
\(\textbf{ARIMA}\) &6.842&6.547&6.464 &2.235&2.014 &1.968&0.839 &0.852 &0.856 \\
\(\textbf{GRU}\) & 5.522 &4.714 &4.588 &2.282 &1.737 &1.632&0.895 &0.923 &0.927 \\
\(\textbf{ST-Net}\) &6.251 &5.045 &4.934 &1.985 &1.721 &1.703&0.865 &0.912 &0.916 \\
\(\textbf{Fine-tuning}\) & 	5.726&4.677&4.367 &1.766&1.714&1.571&0.887 &0.925 &0.934 \\
\(\textbf{MAML}\) &6.066&5.026& 4.677&1.850 &1.741&1.566&0.873 &0.913 &0.925 \\
\(\textbf{DastNet}\)&4.983&4.465&4.143&1.669&\underline{1.586}&1.440&0.914 &0.931 &0.941 \\
\(\textbf{MGAT}\)&4.942&4.452&\underline{4.102}&1.643&1.575&1.427&0.916 &0.932 &\underline{0.942} \\
\(\textbf{ST-DAAN}\) &5.066&4.513&4.191&1.657&1.616&1.430&0.912 &0.930 &0.939 \\
\(\textbf{MetaST}\) &5.057&4.506&4.184&1.673&1.588&1.445&0.912 &0.930 &0.940 \\
\(\textbf{ST-GFSL}\) & 4.863&4.413&4.188&\underline{1.623}&1.611&1.473&0.918 &0.933 &0.940 \\
\(\textbf{CrossTReS}\) &\underline{4.772}&\underline{4.355}&4.147&1.644 &1.601 &1.497&\underline{0.921} &\underline{0.935} &0.941 \\
\(\textbf{TransGTR}\)& 5.123&4.486&4.235&1.671&1.637&1.506&0.910 &0.931 &0.938 \\
\(\textbf{Ada-STGCN}\)&4.922&4.432&4.159&1.651&1.582&1.445&0.916 &0.932 &0.940 \\
\(\textbf{CityTrans}\)&4.960&4.459&4.168&1.653&1.589&1.432&0.915 &0.931 &0.940 \\
\hline
\(\textbf{SFMGTL}\)&\textbf{4.732} &\textbf{4.334} &\textbf{4.019} &\textbf{1.605} &\textbf{1.555} &\textbf{1.393}&\textbf{0.923}&\textbf{0.935}&\textbf{0.944} \\
\hline
\(\textbf{SFMGTL-Full}\)& \multicolumn{3}{c|}{\text{3.245}}&\multicolumn{3}{c|}{\text{1.229}}&\multicolumn{3}{c}{0.964}\\
\hline
\end{tabular}}
\label{Tab:res1}
\end{table}

\begin{table}[htbp]
\scriptsize
\renewcommand\arraystretch{1}
\caption{The experimental results for bike demand prediction}
\centering
\review{\begin{tabular}{c|c|c|c|c|c|c|c|c|c}
\hline
\multirow{3}*{\textbf{Model}}&\multicolumn{9}{c}{\textbf{NY-Bike}$\xrightarrow{}$ \textbf{DC-Bike}}\\
\cline{2-10}
&\multicolumn{3}{c}\textbf{RMSE\quad\quad\quad\quad\quad}&\multicolumn{3}{|c}\textbf{MAE\quad\quad\quad\quad\quad}&\multicolumn{3}{|c}\textbf{$\text{R}^2$\quad\quad\quad\quad\quad}\\
\cline{2-10}
&\textbf{1 Day}&\textbf{3 Day}&\textbf{7 Day}&\textbf{1 Day}&\textbf{3 Day}&\textbf{7 Day}&\textbf{1 Day}&\textbf{3 Day}&\textbf{7 Day}\\
\hline
\(\textbf{ARIMA}\)&4.021&3.845&3.713&1.715&1.632&1.504&0.467 &0.513 &0.546 \\
\(\textbf{GRU}\) &3.017&2.928&2.892&1.447&1.341&1.270&0.700 &0.718 &0.725 \\
\(\textbf{ST-Net}\) &2.828&2.544 &2.501& 1.247&1.196&1.194&0.737 &0.787 &0.794 \\
\(\textbf{Fine-tuning}\) &2.603&2.477&2.390&1.106&1.160&1.091&0.777 &0.798 &0.812 \\
\(\textbf{MAML}\) &2.596&2.475&2.397&1.120&1.177&1.091&0.778 &0.798 &0.811 \\
\(\textbf{DastNet}\) &2.539&2.431&2.324&\underline{1.067}&1.062&1.042&0.788 &0.805 &0.822 \\
\(\textbf{MGAT}\)&\underline{2.524}&2.432&\underline{2.312}&1.091&1.049&1.037&\underline{0.790} &0.805 &\underline{0.824} \\
\(\textbf{ST-DAAN}\) &2.534&2.445&2.335&1.102&1.091&1.067&0.788 &0.803 &0.820 \\
\(\textbf{MetaST}\) &2.541&2.450&2.345&1.080&1.071&1.039&0.787 &0.802 &0.819 \\
\(\textbf{ST-GFSL}\) & 2.534&2.427&2.320&1.074&1.056&1.024&0.788 &0.806 &0.823 \\
\(\textbf{CrossTReS}\)&2.576&\underline{2.425}&2.322&\textbf{1.067}&\underline{1.041}&	\textbf{1.017}&0.781 &\underline{0.806} &0.822 \\
\(\textbf{TransGTR}\)& 2.551&2.428&2.330&1.073&1.054&1.038&0.786 &0.806 &0.821 \\
\(\textbf{Ada-STGCN}\)& 2.571&2.459&2.372&1.087&1.062&1.069&0.782 &0.801 &0.815 \\
\(\textbf{CityTrans}\)& 2.562&2.424&2.338&1.078&1.058&1.035&0.784 &0.806 &0.820 \\
\hline
\(\textbf{SFMGTL}\)&\textbf{2.519} &\textbf{2.421}&\textbf{2.310} &1.069 &\textbf{1.032}&\underline{1.023}&\textbf{0.791} 	&\textbf{0.807} &\textbf{0.824} \\
\hline
\multirow{3}*{\textbf{Model}}&\multicolumn{9}{c}{\textbf{CHI-Bike}$\xrightarrow{}$\textbf{DC-Bike}}\\
\cline{2-10}
&\multicolumn{3}{c}\textbf{RMSE\quad\quad\quad\quad\quad}&\multicolumn{3}{|c}\textbf{MAE\quad\quad\quad\quad\quad}&\multicolumn{3}{|c}\textbf{$\text{R}^2$\quad\quad\quad\quad\quad}\\
\cline{2-10}
&\textbf{1 Day}&\textbf{3 Day}&\textbf{7 Day}&\textbf{1 Day}&\textbf{3 Day}&\textbf{7 Day}&\textbf{1 Day}&\textbf{3 Day}&\textbf{7 Day}\\
\hline
\(\textbf{ARIMA}\)&4.021&3.845&3.713&1.715&1.632&1.504&0.467 &0.513 &0.546 \\
\(\textbf{GRU}\) &3.017&2.928&2.892&1.447&1.341&1.270&0.700 &0.718 &0.725 \\
\(\textbf{ST-Net}\) &2.828&2.544 &2.501& 1.247&1.196&1.194&0.737 &0.787 &0.794 \\
\(\textbf{Fine-tuning}\) &2.616&2.600&2.436&1.096&1.178& 1.079&0.775 &0.777 &0.805 \\
\(\textbf{MAML}\) &2.615&2.602&2.438& 1.095&1.163&1.084&0.775 &0.777 &0.804 \\
\(\textbf{DastNet}\) &2.491&2.456&2.350&1.077&1.107& \underline{1.028}&0.796 &0.801 &0.818 \\
\(\textbf{MGAT}\)&\underline{2.473}&2.451&2.362&1.072&1.101&1.031&\underline{0.799} &0.802 &0.816  \\
\(\textbf{ST-DAAN}\) &2.510&2.478&2.373&1.102&1.118&1.059&0.792 &0.798 &0.815 \\
\(\textbf{MetaST}\) &2.507&2.497&2.373&1.069&1.107&1.040&0.793 &0.795 &0.815 \\
\(\textbf{ST-GFSL}\) & 2.481&\underline{2.431}&\underline{2.345}&\underline{1.068}&1.105& 1.047&0.797 &\underline{0.805} &\underline{0.819} \\
\(\textbf{CrossTReS}\)&2.507&2.468&2.352 &1.081&\underline{1.065}&1.035&0.793 &0.799 &0.818 \\
\(\textbf{TransGTR}\)& 2.494&2.461&2.375&1.093&1.100&1.050&0.795 &0.801 &0.814 \\
\(\textbf{Ada-STGCN}\)&2.552&2.502&2.373&1.094&1.134&1.079&0.785 &0.794 &0.815  \\
\(\textbf{CityTrans}\)&2.539&2.509&2.383&1.069&1.143&1.063&0.788 &0.793 &0.811  \\
\hline
\(\textbf{SFMGTL}\)&\textbf{2.446}&\textbf{2.416}&\textbf{2.343} &\textbf{1.065}&\textbf{1.055}&\textbf{1.023}&\textbf{0.803} &\textbf{0.808} &\textbf{0.819} \\
\hline
\(\textbf{SFMGTL-Full}\)&\multicolumn{3}{c|}{2.032}&\multicolumn{3}{c|}{0.880}&\multicolumn{3}{c}{0.864} \\
\hline
\end{tabular}}
\label{Tab:res2}
\end{table}

\begin{table}[htbp]
\scriptsize
\renewcommand\arraystretch{1}
\caption{Model parameters and performance comparison}
\centering
\review{\begin{tabular}{c|c|c|c|c|c|c|c}
\hline
\multirow{2}*{\textbf{Model}}&\multirow{2}*{\textbf{\# Param}}&\multicolumn{3}{c|}{\textbf{Taxi}}&\multicolumn{3}{c}{\textbf{Bike}}\\
\cline{3-8}
&&\textbf{RMSE}&\textbf{MAE}&\textbf{R}$^2$&\textbf{RMSE}&\textbf{MAE}&\textbf{R}$^2$\\
\hline
\(\textbf{ARIMA}\)&- &33.49\%&26.80\%&9.80\%&37.57\%&35.28\%&59.31\%\\
\(\textbf{GRU}\) &2.5 K&10.76\%&18.35\%&1.86\%&18.22\%&22.63\%&13.14\%\\
\(\textbf{ST-Net}\) &322.6 K&18.25\%&15.75\%&3.86\%&8.04\%&13.84\%&4.67\%\\
\(\textbf{Fine-tuning}\) &322.6 K&8.76\%&7.14\%&1.52\%&4.37\%&6.51\%&2.21\%\\
\(\textbf{MAML}\) &322.6 K&13.76\%&8.98\%&2.74\%&4.38\%&6.79\%&2.21\%\\
\(\textbf{DastNet}\) &213.7 k&2.96\%&2.30\%&0.43\%&0.92\%&1.79\%&0.36\%\\
\(\textbf{MGAT}\)& 246.2 K&2.35\%&1.64\%&0.33\%&0.68\%&1.77\%&0.23\%\\
\(\textbf{ST-DAAN}\) &726.0 K&4.02\%&2.51\%&0.61\%&1.49\%&4.15\%&0.64\%\\
\(\textbf{MetaST}\) &824.5 K&3.85\%&2.48\%&0.58\%&1.75\%&2.15\%&0.77\%\\
\(\textbf{ST-GFSL}\) & 867.4 K&2.06\%&2.44\%&0.27\%&0.56\%&1.66\%&0.18\%\\
\(\textbf{CrossTReS}\)&482.7 K&1.30\%&3.50\%&0.13\%&1.30\%&0.61\%&0.57\%\\
\(\textbf{TransGTR}\)&1381 K&4.10\%&4.24\%&0.60\%&1.25\%&2.19\%&0.52\%\\
\(\textbf{Ada-STGCN}\)&355.2 K&2.45\%&2.14\%&0.33\%&2.51\%&3.93\%&1.17\%\\
\(\textbf{CityTrans}\)& 881.8 K&2.81\%&2.57\%&0.40\%&2.08\%&2.70\%&0.95\%\\
\hline
\(\textbf{SFMGTL}\)&{114.8 K}&-&-&-&-&-&-\\
\hline
\end{tabular}}
\label{Tab:compare}
\end{table}

\subsection{Ablation Study}
To verify the effectiveness of each component of our model, we further implement an ablation study in this section. In detail, we analyze the effects of reconstruction loss, hierarchical node clustering, adversarial training and private meta-knowledge for the taxi demand prediction. 

\textbf{Effects of Reconstruction loss.} To show the necessity of our reconstruction loss, we conduct an experiment where we exclude the corresponding loss term, referred to as SFMGTL w/o RL. The results in Table \ref{Tab:res3} present that the inclusion of the reconstruction loss equips our model to effectively capture static urban structures, and offers notable benefits to the prediction performance to a certain extent. Although we argue in the introduction part that static urban relationships do not always provide useful information for traffic prediction, the finding still underscores the proper integration of static urban structure helps the model learn dynamic relationships.

\textbf{Effects of Hierarchical Node Clustering.} Subsequently, we undertake an experiment wherein the multi-granular information is omitted from the model architecture, referred to as SFMGTL w/o HNC. The experiments indicate a notable degradation in model performance. This observation strongly reflects the rationality of considering multi-granularity during cross-city knowledge transfer since different granularities indicate various scale information. Besides, the training process based on a multi-task paradigm also benefits the integration of various scales.

\textbf{Effects of Adversarial Training.} Different from CrossTReS \citep{jin2022selective} introducing learnable weights for each region, we employ adversarial training to extract domain invariant knowledge. Such components play a crucial role in guaranteeing the domain-invariance of the acquired node embeddings, thereby mitigating negative knowledge transfer. The results reflect the indispensable contribution of adversarial training with a performance drop in SFMGTL w/o AT.

\textbf{Effects of Private Meta-Knowledge.}
Furthermore, to extract knowledge that is specific to the target domain, we incorporate private memory during the training process. To verify the effectiveness of this approach, we conduct a comparison between model performances with and without the inclusion of private memory. Notably, upon excluding the private memory component, a marginal reduction in model performance is observed in SFMGTL w/o PMT. This finding shows that the common memory can actually capture primary useful knowledge among cities, and our model does not benefit a lot from the introduction of the learnable private memory.

\begin{table}[h]
\scriptsize
\caption{Ablation study for taxi pickup/dropoff prediction}
\centering
\review{
\begin{tabular}{c|c|c|c|c|c|c|c|c|c}
\hline
\multirow{3}*{\textbf{Model}}&\multicolumn{9}{c}{\textbf{NY-Taxi}$\xrightarrow{}$ \textbf{DC-Taxi}}\\
\cline{2-10}
&\multicolumn{3}{c|}{\textbf{RMSE}}&\multicolumn{3}{c|}{\textbf{MAE}}&\multicolumn{3}{c}{\textbf{R$^2$}}\\
\cline{2-10}
&\textbf{1 Day}&\textbf{3 Day}&\textbf{7 Day}&\textbf{1 Day}&\textbf{3 Day}&\textbf{7 Day}&\textbf{1 Day}&\textbf{3 Day}&\textbf{7 Day}\\
\hline
\(\textbf{SFMGTL w/o RL}\)&5.066&4.382 &4.090 &1.638 &1.563 &1.485&0.912 &0.934 &0.942\\
\(\textbf{SFMGTL w/o HNC}\)&5.142&4.381 &4.101 &1.733 &1.548 &1.485&0.909 &0.934 &0.942 \\
\(\textbf{SFMGTL w/o AT}\)&5.097&4.395&4.081&1.754&1.550&1.478&0.910 &0.933 &0.943 \\
\(\textbf{SFMGTL w/o PMT}\)&5.057&4.354&4.053&1.660&1.538&1.434&0.912 &0.935 &0.943 \\
\hline
\(\textbf{SFMGTL}\)&5.003&4.331 &4.025 &1.617 &1.503&1.425&0.914 &0.935 &0.944 \\
\hline
\multirow{3}*{\textbf{Model}}&\multicolumn{9}{c}{\textbf{CHI-Taxi}$\xrightarrow{}$ \textbf{DC-Taxi}}\\
\cline{2-10}
&\multicolumn{3}{c|}{\textbf{RMSE}}&\multicolumn{3}{c|}{\textbf{MAE}}&\multicolumn{3}{c}{\textbf{R$^2$}}\\
\cline{2-10}
&\textbf{1 Day}&\textbf{3 Day}&\textbf{7 Day}&\textbf{1 Day}&\textbf{3 Day}&\textbf{7 Day}&\textbf{1 Day}&\textbf{3 Day}&\textbf{7 Day}\\
\hline
\(\textbf{SFMGTL w/o RL}\)&4.777&4.375&4.082&1.658&1.605&1.451&0.921 &0.934 &0.943 \\
\(\textbf{SFMGTL w/o HNC}\)&4.800&4.399&4.085&1.667&1.611&1.470&0.921 &0.933 &0.942 \\
\(\textbf{SFMGTL w/o AT}\)&4.788&4.390&4.081&1.664&1.600&1.457&0.921 &0.934 &0.943 \\
\(\textbf{SFMGTL w/o PMT}\)&4.753&4.361&4.025&1.632&1.559&1.404&0.922 &0.934 &0.944 \\
\hline
\(\textbf{SFMGTL}\)&4.732&4.334&4.019&1.605&1.555&1.393&0.923 &0.935 &0.944 \\
\hline
\end{tabular}}
\label{Tab:res3}
\end{table}

\subsection{Hyperparameter Study} 
\begin{figure}[htbp]
    \centering    
    \includegraphics[width=12cm]{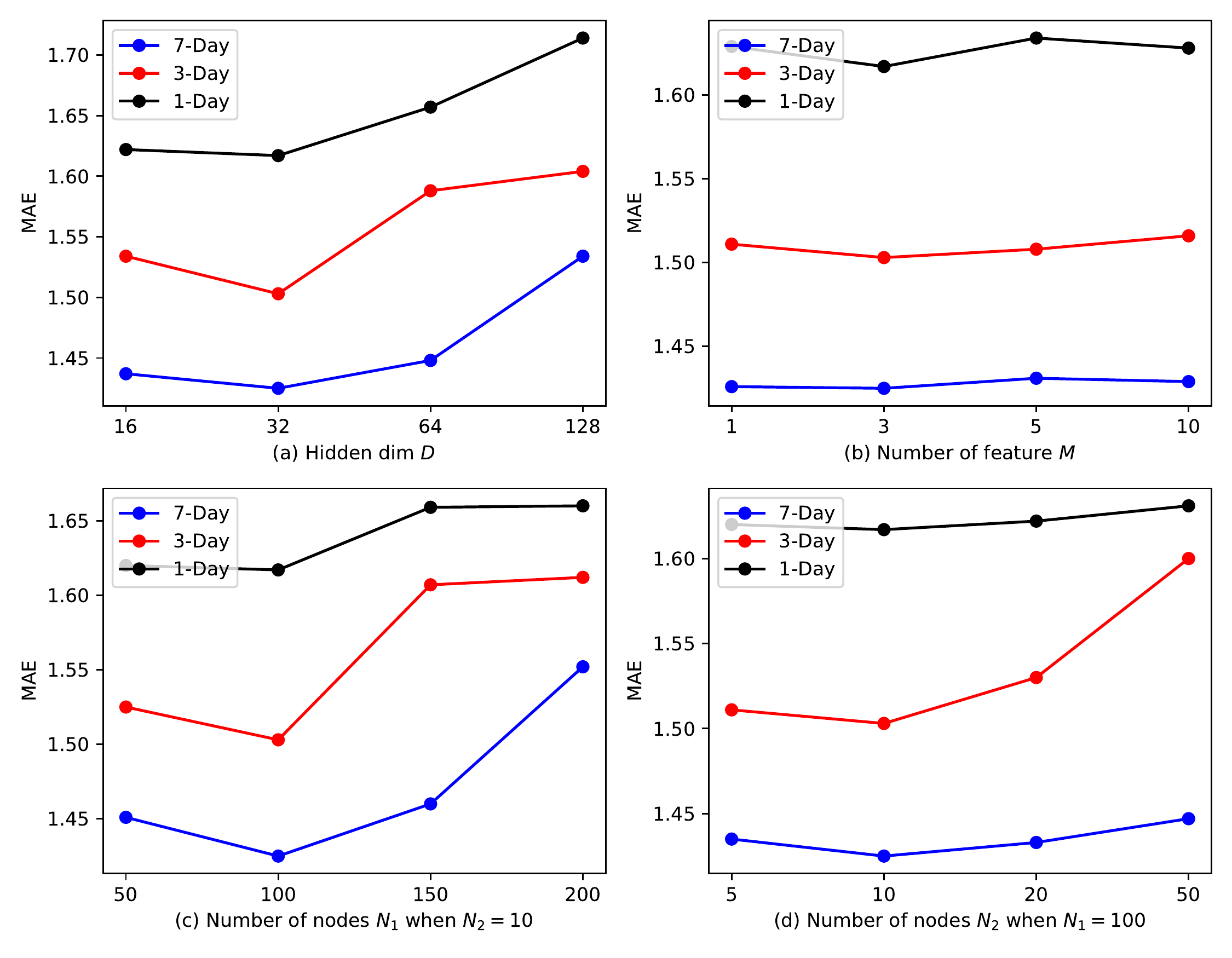}
    \caption{Effects of hyperparameters on MAE.}
     \label{fig:hyperparameter}
\end{figure}

We conducted an exploration of the impact of three key hyperparameters: hidden dim $D$, number of features $M$ in memory modules; and number of nodes $N_1, N_2$ at two levels. For the sake of simplicity, we present the results through an example comparing NY-Taxi to DC-Taxi. The results of this hyperparameter analysis are depicted in Fig. \ref{fig:hyperparameter}. We can draw the following conclusion through observations: (1) Regarding the hidden dimension $D$, our findings show that the most favorable hidden dimension is approximately 32, as deviating from this range tends to induce either underfitting (with fewer dimensions) or overfitting (with higher dimensions); (2) The model's performance demonstrates minimal sensitivity to the number of features $M$ as Fig. \ref{fig:hyperparameter} (b) shows. This suggests that the number of features in the memory modules, within reasonable bounds, does not significantly impact the model's predictive capabilities, making it a less critical hyperparameter to fine-tune; (3) The model's effectiveness is notably influenced by the counts of nodes at each granularity level. Inadequate nodes hinder the exclusive clustering process during training, while excessive nodes fail to capture meaningful zonal information. Therefore, it is vital to choose the clustering nodes meticulously. 

\review{
\subsection{Impact of Source Dataset}
The quality of source datasets plays a crucial role in the knowledge transfer process. Inadequate or poor-quality source datasets may result in negative transfer effects \citep{wang2019characterizing}, potentially diminishing the performance on target cities. To explore the influence of source datasets, we conduct additional experiments by reducing their sizes and introducing noises. 
Specifically, we decrease the size of the source datasets to 1 month, 3 months, and 6 months, and introduce Gaussian noise with standard deviation (SD) of 1 and 5 respectively.

As Table \ref{Tab:res4} presents, even a short duration of the source data benefits the prediction in the target domain, with improvements increasing as the amount of source data grows. Interestingly, we observe that minor disturbances in the source datasets can lead to a slight decrease in RMSE but an increase in MAE in domain prediction. As RMSE can be easily influenced by extreme values and MAE can more evenly depict the average performance, we conjecture the noisy source data increases data variability, prompting the model to predict extreme values more actively. However, the added noise still negatively impacts overall average performance. On the other hand, significant noise (i.e. SD=5) can indeed disrupt useful information within the source domain, resulting in negative knowledge transfer effects.

\begin{table}[h]
\scriptsize
\caption{Taxi pickup/dropoff prediction based on different quality of source datasets}
\centering
\review{
\begin{tabular}{c|c|c|c|c|c|c|c|c|c}
\hline
\multirow{3}*{\textbf{Model}}&\multicolumn{9}{c}{\textbf{NY-Taxi}$\xrightarrow{}$ \textbf{DC-Taxi}}\\
\cline{2-10}
&\multicolumn{3}{c|}{\textbf{RMSE}}&\multicolumn{3}{c|}{\textbf{MAE}}&\multicolumn{3}{c}{\textbf{R$^2$}}\\
\cline{2-10}
&\textbf{1 Day}&\textbf{3 Day}&\textbf{7 Day}&\textbf{1 Day}&\textbf{3 Day}&\textbf{7 Day}&\textbf{1 Day}&\textbf{3 Day}&\textbf{7 Day}\\
\hline
\(\textbf{1-month source}\)&5.734&4.696&4.243&2.614&1.710  &1.487 &0.887&0.924&0.938 \\
\(\textbf{3-month source}\)&5.463 &4.418& 4.141&2.229&1.665  &1.448  &0.897&0.933&0.941 	 	 \\
\(\textbf{6-month source}\)&5.110&4.381& 4.076&2.012&1.606  &1.430 &0.910&0.934&0.943 	 	 \\
\(\textbf{Noisy source with SD= 1}\)&4.917&4.183& 3.874&1.880&1.618   &1.507  &0.917&0.940&0.948  \\
\(\textbf{Noisy source with SD= 5}\)&8.095&6.625& 6.146&5.151&4.650   &4.489  &0.774&0.849&0.870 \\
\hline
\(\textbf{Original dataset}\)&5.003&4.331 &4.025 &1.617 &1.503&1.425&0.914 &0.935 &0.944 \\
\hline
\multirow{3}*{\textbf{Model}}&\multicolumn{9}{c}{\textbf{CHI-Taxi}$\xrightarrow{}$ \textbf{DC-Taxi}}\\
\cline{2-10}
&\multicolumn{3}{c|}{\textbf{RMSE}}&\multicolumn{3}{c|}{\textbf{MAE}}&\multicolumn{3}{c}{\textbf{R$^2$}}\\
\cline{2-10}
&\textbf{1 Day}&\textbf{3 Day}&\textbf{7 Day}&\textbf{1 Day}&\textbf{3 Day}&\textbf{7 Day}&\textbf{1 Day}&\textbf{3 Day}&\textbf{7 Day}\\
\hline
\(\textbf{1-month source}\)& 5.537&4.661&4.212 &2.229&1.798&1.540 &0.894&0.925&0.939 	 	 \\
\(\textbf{3-month source}\)& 5.439&4.428&4.113 & 2.130&1.607 &1.432 &0.898&0.932&0.942 	 	 \\
\(\textbf{6-month source}\)& 5.174&4.399&4.096&1.993&1.587   &1.404  &0.908&0.933&0.942\\
\(\textbf{Noisy source with SD= 1}\)&5.091&4.358& 3.811&1.990&1.806   &1.530  &0.911&0.935&0.950 \\
\(\textbf{Noisy source with SD= 5}\)&7.152&6.740& 6.179&5.093&4.692   &4.506 &0.824&0.843&0.868  \\
\hline
\(\textbf{Original dataset}\)&4.732&4.334&4.019&1.605&1.555&1.393&0.923 &0.935 &0.944 \\
\hline
\end{tabular}}
\label{Tab:res4}
\end{table}
}

\subsection{Case Study}
\begin{figure}[!t]
    \centering    
    \includegraphics[width=17cm]{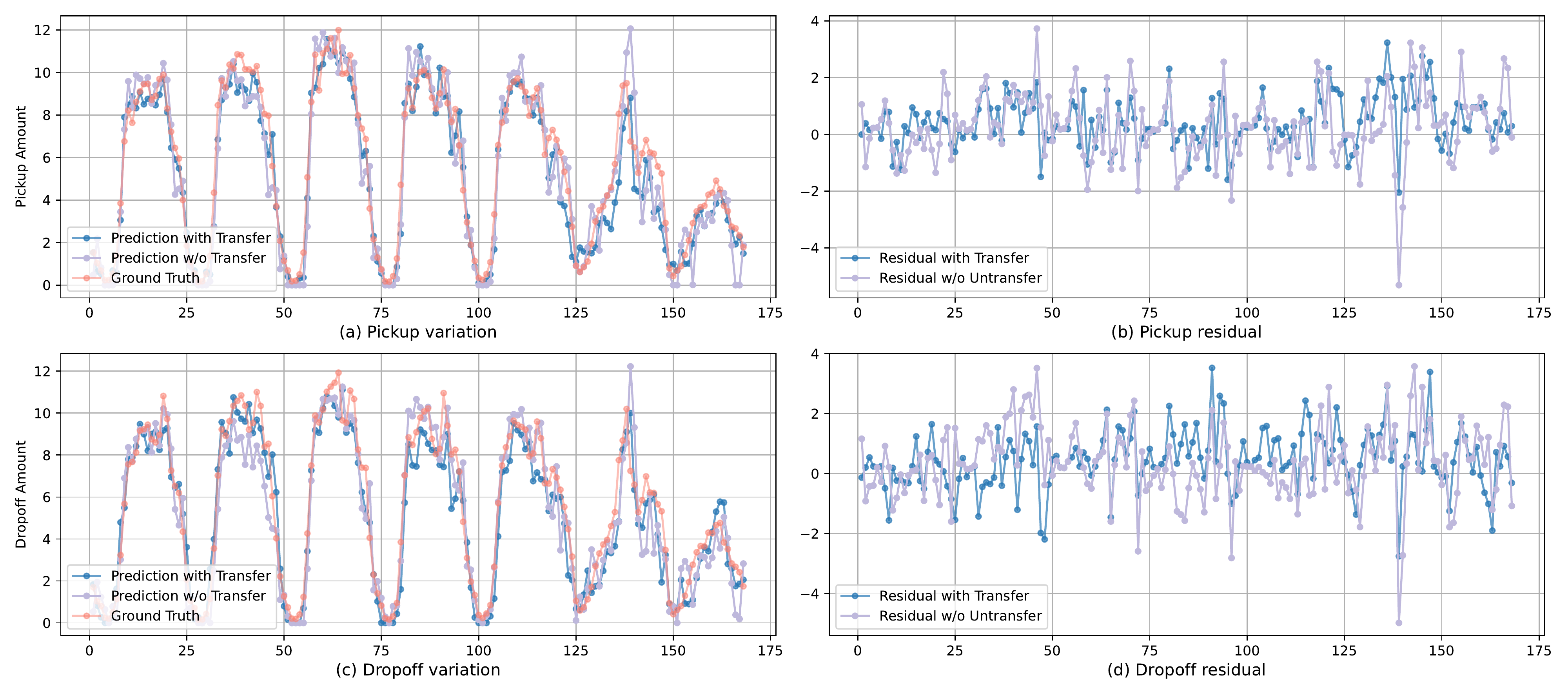}
    \caption{\review{Temporal variation of taxi pickup/dropoff demands within one week. The ground truth and residuals with and without knowledge transfer are visualized.}}
     \label{fig:case_temporal}
\end{figure}

\begin{figure}[htbp]
    \centering    
    \includegraphics[width=17cm]{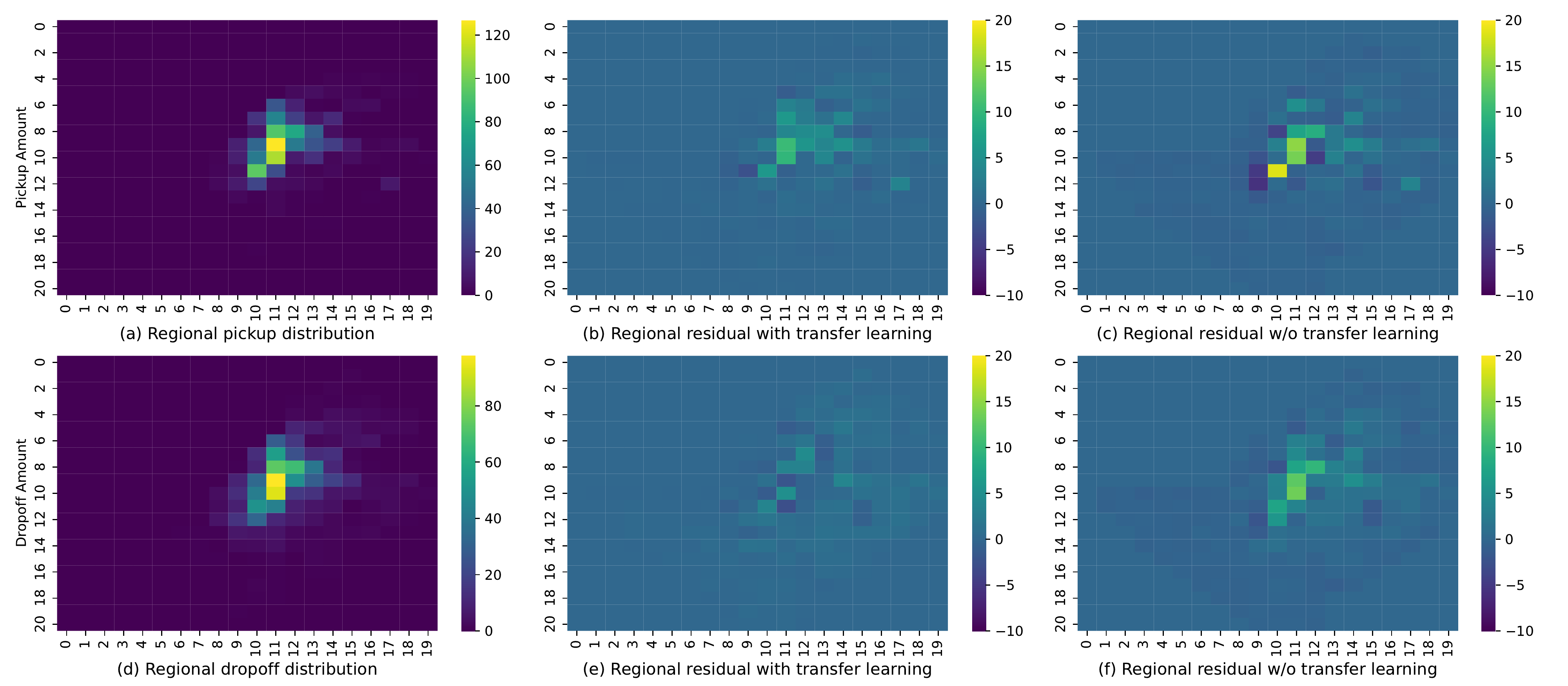}
    \caption{\review{Regional distribution of taxi pickup/dropoff demands within one week. The ground truth and residuals with and without knowledge transfer are visualized.}}
     \label{fig:case_region}
\end{figure}

\review{
In this section, we undertake a case study to offer a comprehensive and intuitive grasp of the prediction results. Specifically, we choose the prediction of taxi demands within one week from Monday to Sunday (i.e. Nov. 7th, 2016 - Nov. 13th, 2016) for demonstration. Afterwards, we compare the performance of our SFMGTL with and without knowledge transfer. 
Fig. \ref{fig:case_temporal} illustrates the comparison of average pickup/dropoff demands across regions among predictions with transfer, predictions without transfer, and the ground truth.  It can be seen that the pickup/dropoff demands have clear periodicity, with noticeable differences in demand variation between weekends and weekdays. Additionally, both models with and without knowledge transfer demonstrate the ability to capture the general variation trends of demands. 
Here, we define the time from 8:00 AM to 8:00 PM as peak time, while all other times are considered off-peak. The MAEs with and without knowledge transfer are 0.868 and 1.114 during peak times, and 0.639 and 0.774 during off-peak time. 
The average improvements of transfer learning during peak and off-peak times are 22\% and 17\%, respectively. On weekends, the improvements can reach 40\% and 36\%, respectively.
Hence, SFMGTL with knowledge transfer exhibits a better enhancement in predicting demands during peak hours, particularly on weekends. 
Conversely, the enhancement in predictions during off-peak times is not as substantial as during peak times.
As depicted in Fig. \ref{fig:case_temporal}, the residual between predictions with knowledge transfer and the ground truth is also lower compared to the prediction through direct training. This observation underscores that transfer learning operates by transferring shared knowledge among cities, particularly during peak hours.

From the regional perspective, Fig. \ref{fig:case_region} presents the regional taxi demands and residuals of predictions. There are total 21$\times$20 regions in DC, and we calculate the average demands for visualization. As the major demands are distributed around the downtown area, transfer learning significantly enhances the prediction accuracy for these regions, resulting in markedly lower residuals compared to non-transferred models.
}

\section{Conclusion}
Data scarcity is ubiquitous in urban areas, especially in low developmental and new districts. To achieve accurate traffic prediction in data-scarce cities, we present a framework known as SFMGTL to address cross-city traffic prediction challenges. In detail, SFMGTL uses semantic fusion modules to learn dynamic graph structures for source and target cities, preserving static urban structures with reconstruction loss. In this way, our model can adapt model structures in various scenarios, and also benefit from long-term urban relationships. In order to extract information from different scales, hierarchical node clustering is employed for granular aggregation, and we train the model at different scales in a multi-task manner. Moreover, we introduce common memories to store shared knowledge, and utilize adversarial training to ensure domain invariance and mitigate negative transfer effects. At last, we also set a private memory to store city-specific knowledge. To show the effectiveness of our model, we conduct experiments on six real-world datasets. \review{Specifically, we predict the taxi and bike demands of different regions in cities. The experimental results demonstrate SFMGTL's superiority over existing methods with lower prediction errors and fewer model parameters. Furthermore, we also implement ablation and hyperparameter studies to present the rationality of each component in the model, and to analyze the robustness of our model. 
To examine the influence of varying quality in source datasets, we truncate the range and introduce noise into the source datasets. The results show that the increase in data sizes facilitates the transfer process. Mild noise encourages the model to adapt to variations actively, whereas strong noise leads to significant negative transfer effects.
Finally, we present a case study to visualize the temporal and regional variations in different scenarios.}

\section*{Acknowledgement}
We would like to acknowledge a grant from RGC Theme-based Research Scheme (TRS) T41-603/20R, and a research grant (project HKUST16207222) from the Hong Kong Research Grants Council under the NSFC/RGC Joint Research Scheme. This study is (partly) supported by the National Natural Science Foundation of China under Grant 52302379. This study is also funded by the Guangzhou Municiple Science and Technology Project (2023A03J0011); and Guangzhou Basic and Applied Basic Research Project (SL2022A03J00083).

\bibliographystyle{model1-num-names}
\biboptions{authoryear}
\bibliography{reference}

\end{document}